# NOVEL AI CAMERA CAMOUFLAGE:
# FACE CLOAKING WITHOUT FULL DISGUISE

**David A. Noever and Forrest G. McKee**
**PeopleTec, Inc., Huntsville, AL**
david.noever@peopletec.com          forrest.mckee@peopletec.com

## ABSTRACT

This study demonstrates a novel approach to facial camouflage that combines targeted cosmetic perturbations and alpha transparency layer manipulation to evade modern facial recognition systems. Unlike previous methods—such as CV dazzle, adversarial patches, and theatrical disguises—this work achieves effective obfuscation through subtle modifications to key-point regions, particularly the brow, nose bridge, and jawline. Empirical testing with Haar cascade classifiers and commercial systems like BetaFaceAPI and Microsoft Bing Visual Search reveals that vertical perturbations near dense facial key points significantly disrupt detection without relying on overt disguises. Additionally, leveraging alpha transparency attacks in PNG images creates a dual-layer effect: faces remain visible to human observers but disappear in machine-readable RGB layers, rendering them unidentifiable during reverse image searches. The results highlight the potential for creating scalable, low-visibility facial obfuscation strategies that balance effectiveness and subtlety, opening pathways for defeating surveillance while maintaining plausible anonymity.

## INTRODUCTION

This paper addresses a novel method to facially disappear using relatively subtle but targeted cosmetic augmentations [1-12]. Nefarious uses of such facial camouflage bank on a wide range of different human motives [13-17] such as concealing a crime or passive scouting, surveilling, or stalking of a target location in a machine-readable world without closed-circuit television (CCTV) detection. Beneficial uses on the other hand support civil libertarians, anti-surveillance laws, identity protection from "doxing", and freedom of anonymous assembly [18-29]. The European laws like GDRP [26] on the "freedom to disappear" on the internet bear particularly for those nations that otherwise restrict or punish such public demonstrators. In the simplest case, one might simply want to own their own facial likeness, meaning that they opt out of participating in mass indexing of portraits and their associate metadata like addresses, phone, or date information [1,8, 20]. Notable law-enforcement users of Clearview.AI and other facial recognition apps [27] are built on scraping social media feeds (>3.5 billion) and have been cited previously as equivalent to putting the entire planet involuntarily into each police case virtually, akin to including anyone in a suspect lineup without opt-in.

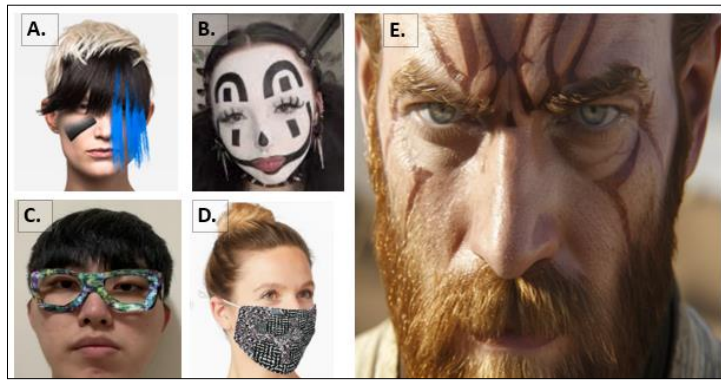

As illustrated in Figure 1, there are several well-known [1-12,17] cosmetic and facial camouflage methods that attempt to hide identities from street surveillance cameras. Traditional makeup techniques can increase the detection confidence inadvertently if they accentuate key points with eye mascara, wrinkle coverage or jaw line shadows. Like the human observer, the computer vision methods based on identifying relative positions between 20-100 facial features (such as eyebrows, nose bridge, lips and jawline) may benefit from prominent eyebrow liners or other focus areas for traditional beauty products. In broad strokes, this work adds to at least three previously found ways to avoid computer vision detectors. One class adds strong skin tonal inverse (black on white, white on black), asymmetries to eye or nose bridge key points, and non-elliptical facial hair boundaries. These methods have been called "CV dazzle" [1, 3-4, 7-8,20] partly in analogy to early WW1 methods to hide ships with checkerboard-patterned dazzle [24, 28-29]. As practical privacy applications, these AI hacks include anti-paparazzi scarves, masks, face painting, anti-CCTV glasses, anti-drone fashion (like head covers) and camera fighting LED lights.

*Figure 1. Approaches to cosmetic surveillance attacks. A. CV dazzle exploits asymmetries: B. Juggalo exploits feature key points like jawlines; C. Adversarial patch attack using glass frame abstract patterns; D. Adversarial patch attack featuring pandemic mask patterns; E. Present subtle skin tone attacks with angular war paint masking eyebrow and jawline key points. The subject in E is generated in MidJourney and does not exist.*

Another class involve decal or fashion attacks that focus the algorithmic search to irrelevant parts of the image with abstract patches or even detailed pictures of umbrellas inside a hand-held photograph [13-14,16-17]. A notable part of this attack class includes its freedom from key points like convolutional neural nets and white or black box adversarial attacks. A final class of anti-detection approaches attack the key points of jaw and cheek detection. Developers call this last approach "Juggalo" makeup after various popular culture makeup trends such as "clown posse" [5-6, 19], the Instagram Juggalo filter, or similar corpse makeup coverage used by metal bands like Kiss and Alice Cooper historically. A common limitation of these previous efforts is their prominence as recognizable disguises. Figure 2 highlights some of the historically aggressive face modifications, all of which would be immediately recognizable by a bystander as theater or disguise. Incidentally, all the facial examples shown in Figure 2 would be recognized by current commercial recognition software [30]. The observer knows the person is hiding their face. In some settings, wearing a mask in public illegal [31-32], even as fashion like hajibs, burqa, or hoodies. Like obscuring license plates from camera detectors, the act of wearing a disguise has the opposite effect of its intentions. Rather than hiding, these camouflage methods draw attention to the wearer. This limitation shares commonality with the "Streisand effect" [33], where the desire for anonymity draws more focus on the subject rather than less.

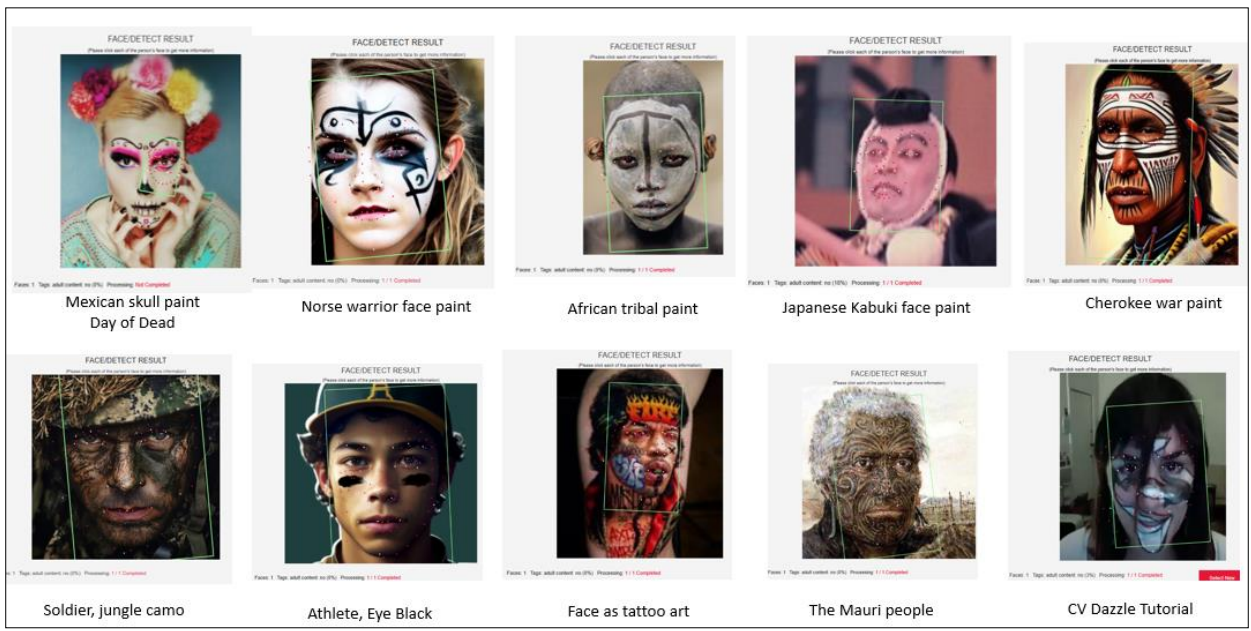

*Figure 2. Failure examples of traditional cosmetic alterations which yield positive face detection*

Figure 2 highlights a further limitation that many obvious, naïve, or historically significant cosmetic changes yield positive face detections. Many of these cases intend to elude either human or machine detectors but even in the case of CV dazzle [4], they all fail to hide from a modern facial recognition system (like [30] betafaceapi.com). There are many other cases where an appropriate disguise—however obvious as a mask—would still play as useful role in avoiding surveillance. For instance, most machine methods require special re-training protocols with 100-1000s of positive examples of cosmetic disguise, and even then, may not have sufficient coverage for the infinite number of minor variations. Similarly if avoiding a trigger detector, like night street CCTV, the attacker may not care if the cosmetic change is attention-drawing, since no face would be detected at all (in some cases) and there is no one other than an algorithm surveying the scene. As recently mentioned on protecting foreign military bases from insurgent attack [6], *"obvious black-and-white clown face patterns would be a giveaway here but consider instead if it was done with the dark greens and blacks and browns more common to military face paint. A subdued design, following the same application of lights and darks and obscuring both jawline, mouth shape and brow line, might be effective in its place."*

This paper explores a novel facial anti-detection class of camouflage that combines a key point attack with more subtle and strategic darkening of skin tone [Figure 1E]. The goal is to investigate if the more effective aspects of previous disguises shown in Figure 1 can persist when their stronger masking elements would not necessarily trigger as a full disguise in a crowd. The paper demonstrates the camouflage empirically using publicly available face detectors first [30], then generates variations on applied cases using random face paint shapes obscuring facial key points. A notable innovation uses a previously described hidden "alpha layer" or transparency attack to stack a grayscale disguise on a positive face detection and make the person disappear in multimodal language models (transformers) but appear unchanged to a human observer [34]. This invisibility cloak effect targets major OpenAI and Microsoft Bing versions of object detection generally and not just faces alone. So for instance, a reverse image search on a face obscured in this way yields no comparable images because the algorithm fails to find the face in the blended image.

**METHODS**

*Generative Face Approach.* The subtle obscuration strategy uses the generative AI (MidJourney [35]), with the following combination prompt to combine two science-fiction characters into an unrecognizable portrait (*Prompt: a photorealistic closeup, a forward-facing portrait image of a white man in late-forties with beard and receding brown hair, Star Wars: Revenge of the Sith movie SONY photorealistic, Darth Maul looming, Obi Wan with lightsaber, classic sci-fi fantasy style*). Although its weights are not open-source or publicly auditable, MidJourney [35] does provide a highly realistic facial generator for natural disguise experiments. Midjourney's core technology appears to be built on an advanced implementation of the latent diffusion model paradigm, distinguished by several proprietary innovations in its architecture. The system employs a modified transformer-based approach for prompt processing, coupled with sophisticated cross-attention mechanisms that enable precise mapping between textual concepts and visual elements. Its architecture reportedly uses multiple specialized sub-models working in concert, with dedicated modules handling composition, style consistency, and detail enhancement within a unified pipeline. The training methodology incorporates elements of contrastive learning and custom loss functions optimized specifically for aesthetic quality, while dynamic attention mechanisms allow adaptive adjustment of prompt elements during generation. These components operate within a highly optimized parallelization framework that enables efficient scaling across GPU clusters, though the specific implementations remain proprietary and continue to evolve with each version release. We employ it here as a hybrid method to generate plausible AI faces that have custom and subtle adversarial elements designed to erode traditional facial recognition.

*Iterative Face Obscuration Approach.* To test the generality of perturbing facial coverage systematically and generate new disguises, this work evaluated programmatically the robustness of Haar cascade-based face detection [36] by introducing randomized visual perturbations over detected facial regions. The process in Table 1 begins by loading an input image and converting it to grayscale. The Haar cascade classifier is applied to detect facial regions, returning the bounding box coordinates (x, y, width, height) and additional detection metadata like neighbors and weights. Once a face is detected, a Shape Manager is initialized with the dimensions and position of the detected facial bounding box. The Random Shape Generator creates various random

```
INPUT: Original image (path)

1. LOAD image as BGR
   CONVERT image to grayscale

2. APPLY Haar cascade face detection:
     DETECT faces with bounding boxes (x, y, width, height)
     STORE face detection results: coordinates, neighbors, weights

3. FOR each detected face:
     INITIALIZE ShapeManager with face bounding box dimensions and offset
     FOR i = 1 to 15:
         GENERATE random shape (rectangle, circle, triangle, or line)
         ADD shape to ShapeManager
     END FOR

4. DRAW all shapes on a copy of the original image (perturbed image)

5. DISPLAY perturbed image

OUTPUT: Perturbed image with randomized obfuscation over detected facial region
```

**Table 1. Pseudocode representation of facial coverage as disguise test framework**

shapes—rectangles, circles, lines, and triangles—with randomized attributes such as position, size, color, thickness, and fill state. These shapes are constrained to the detected face region using offsets. The Shape Manager iteratively adds a predefined number of shapes (e.g., 15), which are then drawn on a copy of the original image, obscuring the facial features. The perturbed image is finally displayed, showing the effect of randomized shapes over the detected face. This approach systematically introduces obfuscation to evaluate whether the Haar cascade can still reliably detect the face, providing insights into its robustness against visual noise. This random coverage spawns millions of alternative facial patterns to test the hypothesis of obscuring key points around the jaw, nose bridge and eyebrow lines. The original face receives a different cosmetic obfuscation on every iteration and the collection of those patterns that pass a Haar detection get saved and analyzed for common elements.

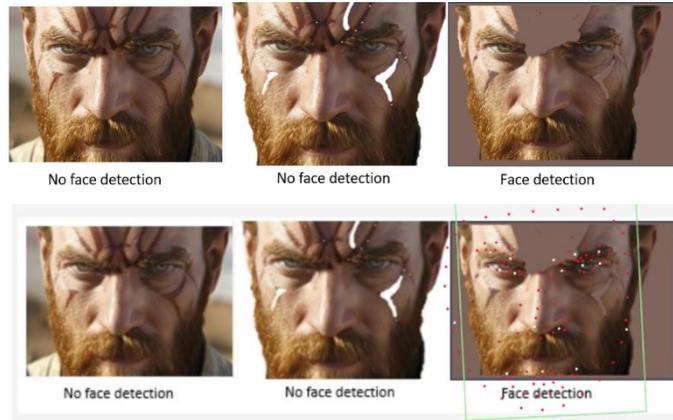

*Figure 3. Ablation study of key points with subtle disguise and brow line focus points*

**RESULTS**

The major finding of this empirical investigation was identification of a class of low-level cosmetic alterations that could influence facial recognition. Figure 3 shows a series of ablations from the baseline image of a more subtle cosmetic alteration than the three previous disguise classes: Juggalo, CV dazzle or adversarial patches. By removing the face pain line by line, the results show the strong influence of the eyebrow line, particularly in the absence of a visible jawline because of the cropping and beard in the subject. Only when the vertical lines lose focus does the modern facial recognition software gain recognition again as a face of any kind. Removal of the cheek paint does not alter the disguise. The detection algorithm [30] suggests that 22 basic facial points (including intra-point distances) make up the positive face detection. In previous work [37], more than 60 key points defined the basic emotional detection, all of which highlighted the jaw, brow, nose bridge, lips, and eyes. One method for future investigation therefore suggests that the skin tonal makeup that perturbs these highly dense facial regions may determine the success of facial recognition generally.

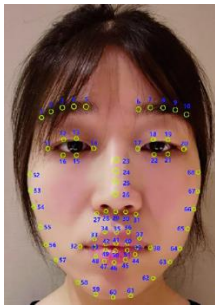

*Figure 4. Key points in facial recognition center [37] on lips, eyes, jawlines, and nose-bridge, while sparsely representing forehead, hair, or cheek regions.*

Given one image that when ablated of cosmetics becomes recognizable and a similar manipulated image that is unrecognizable, one can imagine a method to blend them and test against more advanced vision-language models. Previous work [34] has demonstrated that face swapping can occur in advanced models when a human-visible alpha layer and a machine-visible red-green-blue (RGB) layer contradict. In Supplemental Material 1, that work showed that computer vision for PNG and other layered formats with transparency present deceptive challenges when a large transformer flattens the alpha layer pixels in favor of a standard RGB matrix of pixels. To explore this sandwiched attack experimentally, a top layer (alpha) of the transparent PNG format was visible from Figure 4 as obviously a face. But the computer vision recognition software fails to identify the RGB layer of the disguised face (otherwise invisible to the human eye). The adversarial

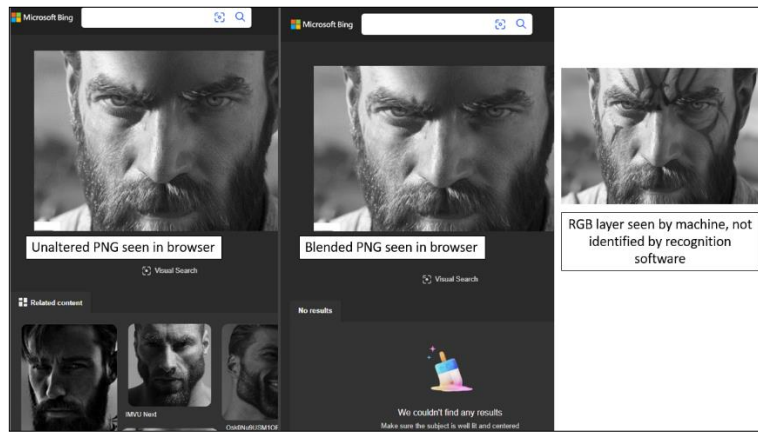

*Figure 5. Transparency attack that sandwiches a cosmetically optimized image, so the Microsoft Bing algorithm fails to find the face contours in grayscale PNG*

sequence shown in Figure 5 illustrates the effects on Microsoft Bing (Visual Search) where the unaltered PNG yields related content (similar faces) but the altered PNG produces no matches because of the under layer RGB seen by the Microsoft algorithm fails to find the face with subtle makeup. This transparency attack implies that a determined actor could effectively poison large facial recognition datasets with PNG samples where the human-visible (alpha) layer always showed the face, but the machine-visible (RGB) layer obscured the detectability or identity.

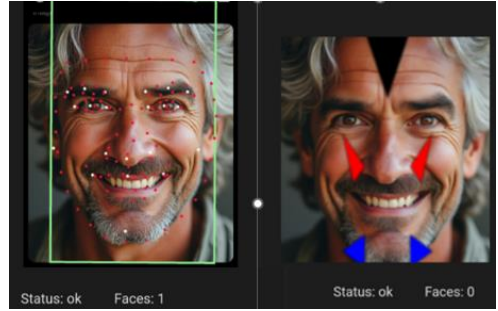

The ability to create viable subtle disguises offers new opportunities to defeat surveillance systems or identity trackers without resorting to obvious makeup tricks. To examine whether the disguise shares some similar origins to the clown posse or Juggalo methods of distorting the brow and jawlines, this work generated random patterns and looked for disguise patterns in face detectors like Haar cascades or alternative online vision-language transformers. Figure 6 shows an example attack against the BetaFaceAPI, where prominent angular distortions of the forehead, mouth and jawline render no-face detected. This finding supports the hypothesis that subtle alterations can defeat key point detections and highlight the potential dense regions of standard distance databases that might identify not just the face but also the personal identity (e.g. Clearview AI names [27]). While this makeup would draw attention to a person in a crowd, the simplicity of disrupting the jaw and brow suggests more covert disguises from purely mechanical surveillance without any onlookers such as CCTV or passive observation or unmanned camera sites. Using the full disguise iteration in Table 1, the transparency of the random shapes can vary the expected success of the disguise (Supplemental Material 2). Facial tattoos generally do not disguise the facial recognition (Supplemental Material 3).

*Figure 6. Key points (left) in positive face detection and angular overlays distorting dense key point areas around the mouth, jawline and forehead rendering no face detected. BetaFaceAPI results shown.*

To highlight the disguise in a side-by-side way, pairwise image were submitted to test the disguise effectiveness (Figure 7). It is worth noting that both the transparency PNG attack and the skin tonal war point effects are transferable to other facial recognition platforms including Microsoft Bing (Figure 5), PimEyes Reverse Face Search [38].

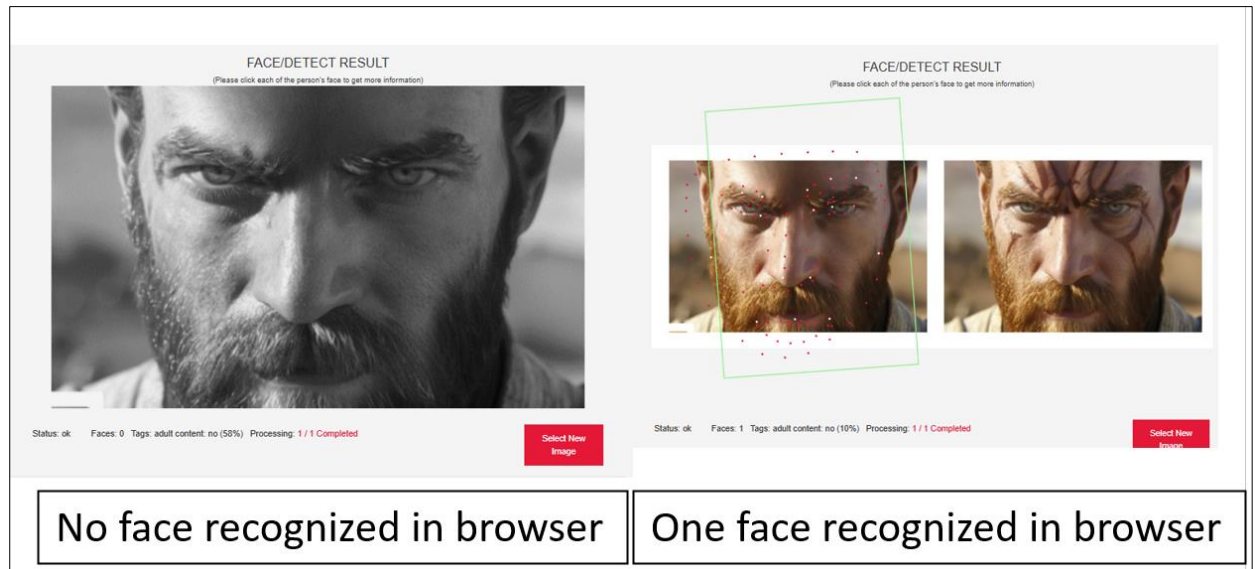

*Figure 7. A. Fooling BetaFaceAPI (left) with a Transparent Layer (visible) and RGB Layer with Skin Tonal Cosmetic Coverage. B. Side-By-Side Submission (right) Which Fails to Identify the Cosmetically Altered Image Next to the Baseline Face.*

**DISCUSSION AND PREVIOUS WORK**

The presented work offers an advancement over prior research into camera camouflage and privacy-enhancing techniques by introducing a novel class of subtle and targeted facial obfuscation methods. While previous efforts, such as CV Dazzle, adversarial patches, and Juggalo makeup, relied on bold, high-contrast modifications to disrupt facial detection, these approaches often suffer from two critical limitations: their theatrical prominence makes them easily recognizable to human observers, and they fail to address modern face detectors trained on robust key-point models. In contrast, this study demonstrates that effective disruption of facial recognition can be achieved through subtle darkening of high-density key-point regions (e.g., brow lines, nose bridge, and jaw contours) without triggering the visibility issues inherent to overt disguises. This nuanced approach mitigates the "Streisand effect," where obvious efforts to hide one's identity paradoxically draw attention.

Unlike traditional methods that focus solely on visual obfuscation for machines, this research introduces a multi-layered strategy that manipulates both human and machine perception. By leveraging the alpha transparency layer of PNG images, the study reveals how machine-readable underlayers (e.g., RGB data) can be decoupled from human-visible content. While human observers perceive a clear, unobscured face, the transparency-layered attack effectively erodes the ability of computer vision models—such as Microsoft Bing Visual Search and BetaFaceAPI—to detect or match identities. This dual-layer invisibility cloak represents a distinct innovation compared to previous approaches, which typically failed when subjected to reverse image searches or advanced vision-language models. Furthermore, by integrating generative AI (MidJourney) and systematically testing random cosmetic perturbations via Haar cascade classifiers, this work explores the generative potential for creating millions of subtle disguises that strategically disrupt facial recognition algorithms. Seeking this disguise diversity potentially avoids just retraining existing facial recognition methods on sample disguises, since the unlimited output of shapes, colors and coverage could refresh new alternatives.

Another crucial distinction lies in the systematic iterative obscuration process introduced. Previous research often relied on static disguise templates, such as face paint patterns or optical misdirections, which lacked adaptability and scalability. In contrast, this study evaluates facial key-point vulnerabilities programmatically, iteratively generating random obfuscations to isolate effective patterns that break recognition while minimizing visual disruption. By analyzing the results of these iterations, common elements of successful disguises—such as vertical line perturbations near the brow—are identified and empirically tested. This iterative strategy introduces a dynamic framework for future testing and refining facial obfuscation techniques, ensuring their applicability across diverse datasets and detectors.

While this study introduces advances in subtle facial camouflage, it also has certain limitations. The method primarily focuses on static images, and its effectiveness against video-based recognition systems remains to be tested. Real-time facial recognition systems often leverage temporal consistency, tracking facial motion over frames, which may reduce the efficacy of randomized static perturbations. Additionally, while the alpha transparency attacks demonstrated success against specific computer vision models, these findings may not generalize universally across all facial recognition systems, particularly those trained on adversarially robust datasets or using multi-modal recognition pipelines that combine depth, texture, and contextual features. Another limitation stems from the reliance on Haar cascade classifiers as a primary testing framework; while still widely used, Haar-based methods are relatively outdated compared to deep learning-based detectors such as MTCNN, RetinaFace, or YOLO. Future work would need to assess whether the proposed disguises remain effective against these advanced models. Finally, subtle cosmetic modifications, while less attention-grabbing than traditional disguises, may still invite scrutiny in certain social contexts, especially where facial irregularities could stand out due to cultural or legal norms. Perhaps a crowd event or sports stadium invites dramatic theatrical modifications without drawing attention but still offer relative obscurity to the wearer from camera surveillance or crowd-counting software. These limitations highlight the need for ongoing experimentation across diverse models, environments, and real-world applications to further validate and generalize the findings.

**CONCLUSIONS AND FUTURE WORK**

In summary, this research advances the field of facial camouflage by bridging the gap between effectiveness and subtlety. The disguise attack targets both traditional Haar cascades and much larger vision transformers such as Microsoft Bing and PimEyes facial image searches. Unlike previous work, which either relied on overtly theatrical disguises or failed against modern detectors, this study provides a novel and potentially scalable framework that combines targeted facial perturbations, multi-layered transparency attacks, and generative AI to disrupt machine vision while maintaining human plausibility. By demonstrating successful obfuscation against commercial detection systems and vision-language transformers, this work not only enhances privacy protections but also opens pathways for further research into adversarial camouflage techniques that balance functionality, subtlety, and practical utility.

Several extensions of this research could further refine and expand its applicability. First, future work could incorporate dynamic perturbation methods that adapt in real time to specific facial recognition models or environmental conditions, such as lighting changes or camera angles. For instance, the iPhone facial recognition software reportedly uses depth cameras to recognize faces in the dark and thus would not be subject to this skin-tone camouflage. Integrating adversarial machine learning techniques could generate personalized obfuscation patterns tailored to an individual's unique facial structure, providing greater precision in attacking key-point detections. The main sample image of a Star Wars mixture here may take advantage of the subject's beard to obfuscate some key points around the jawline for instance. Second, exploring multi-spectral disguises that consider infrared or thermal imaging cameras may address detection systems beyond the visible spectrum, particularly in surveillance applications. Third, hybrid methods that combine 3D facial morphing—using subtle depth distortions of facial geometry—with cosmetic obfuscations could further evade structured light or LiDAR-based recognition systems. Additionally, automated optimization frameworks based on generative adversarial networks (GANs) or diffusion models could streamline the creation of randomized disguise patterns, reducing the need for manual iteration while improving disguise success rates. Finally, applying this approach to live video feeds or augmented reality (AR) filters could expand its use cases, such as real-time obfuscation for wearable devices or video conferencing platforms. These areas offer a rich research opportunity within the growing field of camera camouflage and privacy-enhancement without undue crowd attention.

## ACKNOWLEDGEMENTS

The authors thank the PeopleTec Technical Fellows program for research support.

*Supplemental Material I. Alpha Transparency Algorithm*

```
Procedure LoadAndPreprocessImage(path, size):
    Read image from path as grayscale
    Resize image to specified size
    Convert grayscale image to RGB format
    Return preprocessed image

Procedure BlendImages(target_images_dir, background_path, size, steps,
learning_rate):
    background_image <- LoadAndPreprocessImage(background_path, size)
    background_tensor <- ConvertToTensor(background_image) * 0.5

    For each filename in target_images_dir:
        If filename ends with ".jpg":
            target_image <-
LoadAndPreprocessImage(PathJoin(target_images_dir, filename), size)
            target_tensor <- ConvertToTensor(target_image)
            alpha <- InitializeTensorWithOnes(DimensionOf(background_tensor))

            optimizer <- InitializeAdamOptimizer(alpha, learning_rate)
            white_background <-
InitializeTensorWithOnesLike(background_tensor)

            For step from 0 to steps:
                ResetGradients(optimizer)
                blended_image <- alpha * background_tensor + (1 - alpha) *
white_background
                loss <- ComputeMSELoss(blended_image, target_tensor)
                BackpropagateLoss(loss)
                UpdateOptimizer(optimizer)

                If step modulo 100 equals 0:
                    LogStepProgress(filename, step, loss)

            SaveBlendedImage(background_tensor, alpha, target_images_dir,
filename)

    Log("Processing complete.")

Main:
    Define target_images_dir, background_path, size, steps, learning_rate
    BlendImages(target_images_dir, background_path, size, steps,
learning_rate)
```

Table 2. Pseudocode for blending a transparent face layer with a RGB "no-face" layer to disguise the recognition software

Supplemental Material 2. Effect of shape transparency on disguise success. Less transparent coverage (middle) renders the face unrecognized, while increasing the transparency eventually yields sufficient key points to trigger a positive detection and bounding box (BetaFaceAPI [30]).

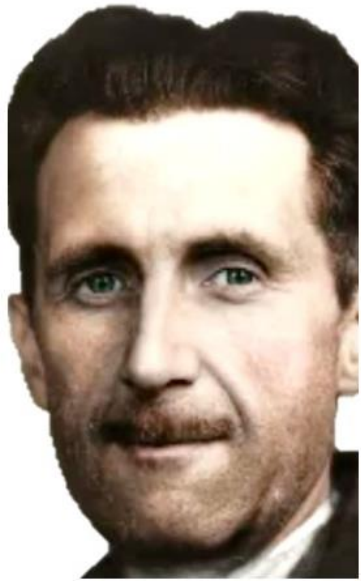 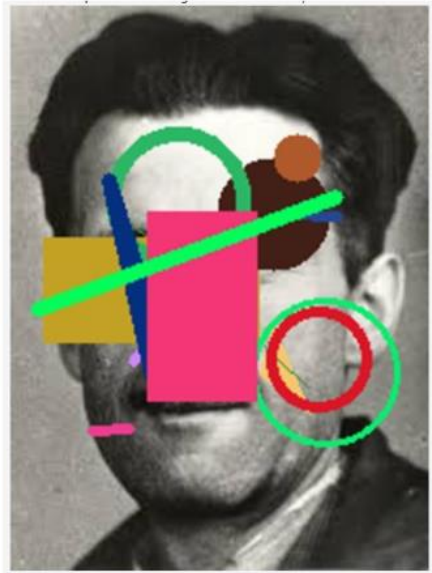 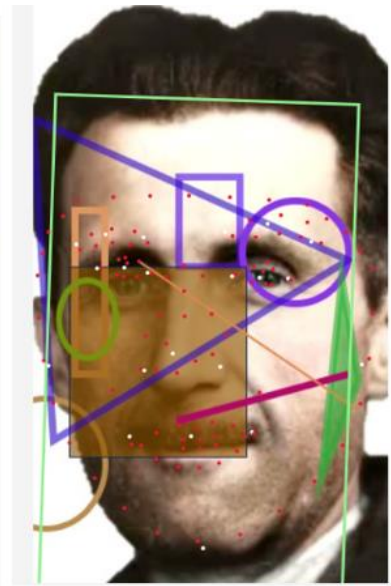

Supplemental Material 3. Facial Tattoos Fail to Disguise the Facial Recognition in All Cases Examined Except for Nearly 100 Percent Coverage with Angular Lines.

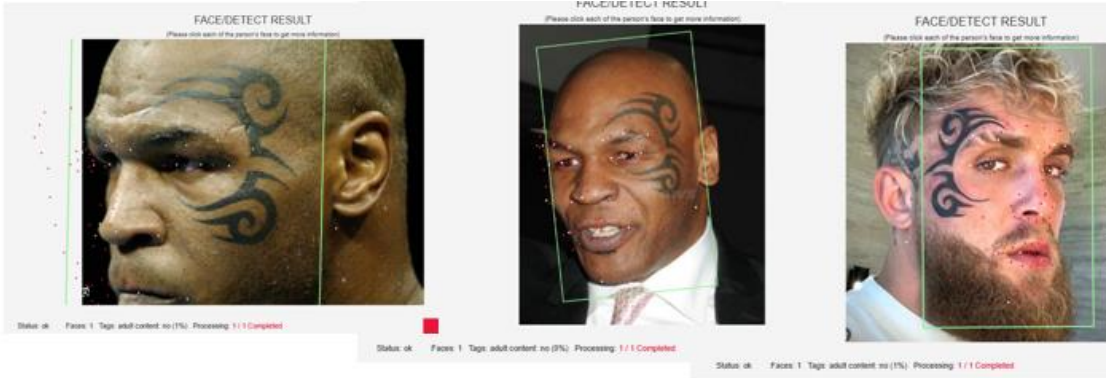

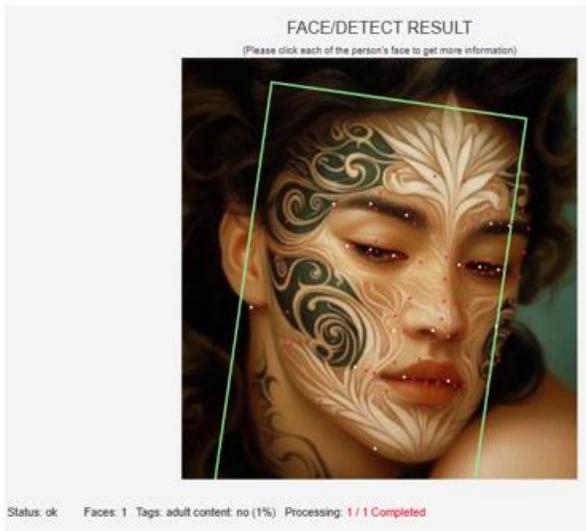
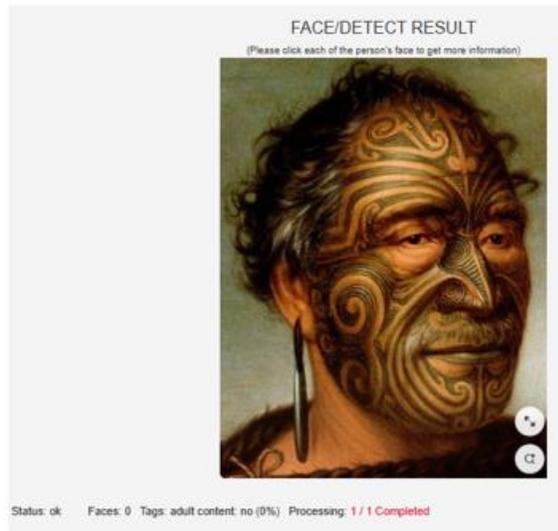